\documentclass{article}

\usepackage{microtype}
\usepackage{graphicx}
\usepackage{subfigure}
\usepackage{booktabs} % for professional tables
\usepackage{hyperref}

\usepackage{algorithm}
\usepackage{algorithmic}
\usepackage{natbib}

\usepackage{arxiv}

\title{Directed Exploration for Reinforcement Learning}

\date{} 					% Or removing it

\author{
  Zhaohan Daniel Guo \\
  Carnegie Mellon University\\
  \texttt{zguo@cs.cmu.edu} \\
   \And
  Emma Brunskill \\
  Stanford University\\
  \texttt{ebrun@cs.stanford.edu} \\
}

\begin{document}
\maketitle

\begin{abstract}
Efficient exploration is necessary to achieve good sample efficiency for reinforcement learning in general. From small, tabular settings such as gridworlds to large, continuous and sparse reward settings such as robotic object manipulation tasks, exploration through adding an uncertainty bonus to the reward function has been shown to be effective when the uncertainty is able to accurately drive exploration towards promising states. However reward bonuses can still be inefficient since they are non-stationary, which means that we must wait for function approximators to catch up and converge again when uncertainties change. We propose the idea of directed exploration, that is learning a goal-conditioned policy where goals are simply other states, and using that to directly try to reach states with large uncertainty. The goal-conditioned policy is independent of uncertainty and is thus stationary. We show in our experiments how directed exploration is more efficient at exploration and more robust to how the uncertainty is computed than adding bonuses to rewards.
\end{abstract}

\section{Introduction}

Neural networks have been shown to be an effective function approximator for large, complex functions in many domains. In order to scale up reinforcement learning, function approximators are a key component, and recent work combining RL with deep learning has shown promising results; algorithms such as DQN, PPO, A3C have been shown to work in domains such as Atari games, and robot locomotion and manipulation \citep{mnih2013playing,schulman2017proximal,mnih2016asynchronous}. However, most approaches only use simple exploration strategies that add some simple randomness to the actions. DQN uses e-greedy, whereas PPO and A3C follow the policy gradient with policies that add some random noise to the actions. The lack of more sophisticated exploration has hindered progress in more complex domains with sparse rewards.

Recently, there has been work towards more sophisticated exploration strategies in deep RL. Noisy networks add e-greedy exploration in the parameter space rather than the action space have been shown to be better \citep{fortunato2017noisy}. Bootstrapping and ensembles have been used to approximate posterior sampling for better exploration \citep{osband2016deep}. Currently, uncertainty-based methods have been shown to be the most effective and most promising to tackle hard exploation domains \citep{houthooft2016vime,ostrovski2017count,tang2017exploration,burda2018large}. These uncertainty-based methods use a reward bonus approach, where they compute a measure of uncertainty and transform that into a bonus that is then added into the reward function. Unfortunately this reward bonus approach has some drawbacks. The main drawback is that reward bonuses may take many, many updates before they propagate and change agent behavior. This is due to two main factors: the first is that function approximation itself needs many updates before converging; the second is that the reward bonuses are non-stationary and change as the agent explores, meaning the function approximator needs to update and converge to a new set of values every time the uncertainties change. This makes it necessary to ensure that uncertainties do not change too quickly, in order to give enough time for the function approximation to catch up and propagate the older changes before needing to catch up to the newer changes. If the reward bonuses change too quickly, or are too noisy, then it becomes possible for the function approximator to prematurely stop propagation of older changes and start trying to match the newer changes, resulting in missed exploration opportunies or even converging to a suboptimal mixture of old and new uncertainties. Non-stationarity has already been a difficult problem for RL in learning a Q-value function, which the DQN algorithm is able to tackle by slowing down the propagation of changes through the use of a target network \citep{mnih2013playing}. These two factors together result in slow adaptation of reward bonuses and lead to less efficient exploration.

We propose the idea of directed exploration, an alternative approach for using uncertainty for exploration, which avoids the issue of non-stationarity altogether. The idea is that instead of transforming the uncertainty into a bonus that is added onto reward, we directly try to visit those states that have high uncertainty. We can learn a goal-conditioned policy $\pi(s,g)$ that would enable us to try to reach any goal states we specify \citep{schaul2015universal,andrychowicz2017hindsight}. We would then repeatedly pick states that have high uncertainty and set them as goals and use the goal-conditioned policy $\pi(s,g)$ to reach them. This results in an algorithm that is completely stationary, because the goal-conditioned policy is independent of the uncertainty. Also, because we can pick and commit to reaching goal states, we are much more robust to the case where uncertainty estimates are changing and noisy, which can negatively impact and slow down reward bonus approaches. We show in our experiments that directed exploration is more robust to noisy and inaccurate uncertainty measures, and is more efficient at exploration than the reward bonus approach.

This idea of directed exploration is closely related to the hierarchical reinforcement learning literature. It can be considered a particular instantiation of goal generation from higher level policies, and then trying to reach those goals through lower level policies. Prior work has focused on uniformly random or expert guided goal generation for learning sub task and task structure, but we look into uncertainty-based goal generation explicitly for better exploration \citep{Held2018AutomaticGG,NIPS2018_7591}. We also do not require learning a hierarhical policy, allowing our method to be more easily adapted to existing algorithms.

\section{Background}

\subsection{Reinforcement Learning}

A basic formulation of the reinforcement learning process is represented by a Markov decision process (MDP). An MDP is a tuple $\langle S,A,T,R,\gamma \rangle$ where 
$S$ is a set of states of the environment, $A$ is a set of actions, $T$ is a transition 
model where $T(s'|s,a)$ is the probability of the environment transitioning to state $s'$ given current state $s$, and the algorithm taking action $a$. $R(s,a)$ is the expected
reward (bounded) received in state $s$ upon taking action $a$, and $\gamma$ is the discount 
factor. 
A policy $\pi$ is a mapping from states to actions. The 
value $V^{\pi}(s)$ of a policy $\pi$ is the expected sum of discounted rewards 
obtained by following $\pi$ starting in state $s$: $V^{\pi}(s) = \sum_{t=1}^{\infty} \gamma^{t-1} r_t $. The optimal policy $\pi^*$ for an MDP is the one with the highest value function, denoted $V^{*}(s)$. In reinforcement learning, the transition and reward models are unknown. A reinforcement learning algorithm attempts interacts with an MDP and tries to learn the optimal or near-optimal policy.

\subsection{DQN}

Deep Q-Networks (DQN) is a deep reinforcement learning algorithm that has been shown to successfully achieve superhuman performance in variety of domains, including multiple Atari games \citep{mnih2013playing}, and is suited to use with discrete action domains. It is a variation of Q-learning and is able to make use of off-policy data, i.e. data that came from following a different policy than the current greedy policy associated with the learned Q-network. This is important as when we are doing directed exploration, we will be following different policies to try to reach different goal states, which will be different from the current best policy.

\subsection{DDPG}

Deep deterministic policy gradient (DDPG) is an off-policy, deep reinforcement learning algorithm to use with continuous state and action spaces based on policy gradient \citep{lillicrap2015continuous}.

\subsection{UVFA}

Universal value function approximators (UVFA) extends the idea of a value function to a goal conditioned value function $V(s, g)$ that represents the value when trying to reach the goal $g$ starting from state $s$ \citep{schaul2015universal}. Given such a value function, we can then navigate to any goal $g$, in particular we can use this to facilitate directed exploration by picking goals $g$ that are useful for exploration. DQN can be naturally extended to implement UVFA by adding the goal $g$ as an additional input to learn the goal-conditioned Q-value function $Q(s, a, g)$, and DDPG can be naturally extended to learn the goal-conditioned policy $\pi(s, g)$. To unify both approaches, since we can extract out a policy from the Q-value function, we now will refer to both DQN and DDPG being able to learn a goal-conditioned policy $\pi(s,g)$.

\subsection{HER}

Hindsight experience replay (HER) is a technique that can speed up the learning of an UVFA by making use of trajectories that fail to reach the goal \citep{andrychowicz2017hindsight}. Given a trajectory that tried and failed to reach $g$, instead we can pretend that we were actually trying to reach $s'$, where $s'$ is a state that we actually visited during the trajectory. HER turns a negative example for a goal $g$ into data for a positive example for a reached goal $s'$.

\section{Directed Exploration}

\subsection{Directed Exploration Outline}

\begin{algorithm}[tb]
\begin{algorithmic}[1]
\STATE \textbf{Input:} $A, U, \pi(s, g)$
\STATE $A$ is an off-policy RL algorithm
\STATE $U$ is a method to compute uncertainty for a state
  \FOR{Episode $i=1$ to $\infty$}
   \WHILE{episode not ended}
        \STATE Pick $g$ with largest uncertainty according to $U$
        \STATE Visit $g$ using goal-conditioned policy $\pi(s, g)$
        \STATE Take a random action
   \ENDWHILE
   \STATE Update $A$
\ENDFOR
\end{algorithmic}
\caption{Generic Directed Exploration RL}
\label{alg:genericdirected}
\end{algorithm}

The general steps behind directed exploration is outlined in Algorithm \ref{alg:genericdirected}. The main idea is to repeatedly try to visit the states with the largest uncertainty according to some uncertainty measure $U$, and then take one step of random action as exploration. We then rely on an off-policy RL algorithm to learn from these directed exploration trajectories. While this outline can be implemented exactly in the tabular setting, there are several additional considerations that must be made when we move to the function approximation setting.

\subsection{Directed Exploration with Function Approximation}

\begin{algorithm}[tb]
\begin{algorithmic}[1]
\STATE \textbf{Input:} $D, K, N, E, A, U$
\STATE $A,E$ are off-policy RL algorithms
\STATE $A$ is used to learn from the real reward
\STATE $E$ is used to learn the UVFA/goal-conditioned policy $\pi(s,g)$
\STATE $U$ is a method to compute uncertainty for a state
\STATE $G \gets$ FIFO buffer of goal states with capacity $N$
  \FOR{Episode $i=1$ to $\infty$}
   \STATE With probability $0.5$, act greedily w.r.t. to current optimal policy
   \STATE Otherwise, do directed exploration:
   \WHILE{episode not ended}
        \STATE Sample goal state $g$ from top $K$ uncertain goal states in $G$
        \STATE Try to reach $g$ for $D$ steps
        \STATE If we reach or $D$ steps are up, then do one step of random action
   \ENDWHILE
   \STATE Store episode experience into common replay buffer
   \STATE Sample minibatch $B$ from common replay buffer
   \STATE Compute uncertainty for each state in $B$ using $U$ and add to $G$
   \STATE Update $A$ with $B$
   \STATE Update $E$ with $B$ using HER
   \STATE Update $U$ with $B$
\ENDFOR
\end{algorithmic}
\caption{Directed Exploration RL with Function Approximation}
\label{alg:directed}
\end{algorithm}

\begin{figure}[tb]
    \centering
    \includegraphics[width=0.4\textwidth]{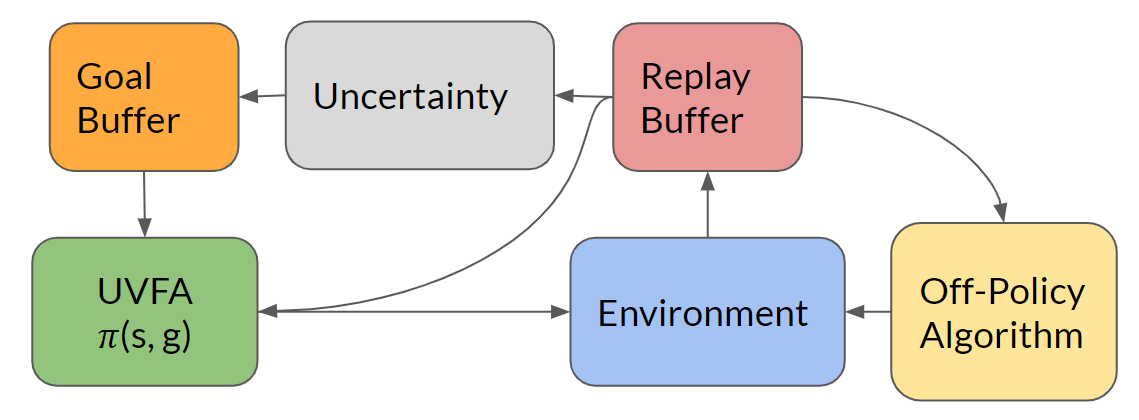}
    \caption{The different components of directed exploration with function approximation.}
    \label{fig:dediagram}
\end{figure}

Algorithm \ref{alg:directed} outlines how to do use directed exploration in the function approximation setting. Figure \ref{fig:dediagram} shows the different components that make up the algorithm.

\subsubsection{Learning a Goal-conditioned Policy}

The first challenge is to get a goal-conditioned policy $\pi(s,g)$ to use for reaching goal states. We use an off-policy RL algorithm $E$ to learn a UVFA/goal-conditioned policy (the green node in Figure \ref{fig:dediagram}) through the use of hindsight experience replay (HER), which is simultaenously being trained along with the existing off-policy RL algorithm $A$ (the yellow node in Figure \ref{fig:dediagram}). Unfortunately, since we are now learning an approximation of $\pi(s,g)$, this means that we may fail to reach our target goal state $g$, and so we introduce a timestep limit $D$. We only follow $\pi(s,g)$ for up to $D$ steps, after which we take one random action step. $E$ is trained from the same common replay buffer as the existing algorithm $A$. In our experiments, we set $D$ to be half or a quarter of the maximum episode length. 

\subsubsection{Tracking Uncertainty and Goal States}

The next challenge is in picking the states with the largest uncertainty. We use a FIFO buffer which stores the states to be used as goal states, as well as their associated uncertainties (orange node in Figure \ref{fig:dediagram}). This buffer is also updated from minibatches sampled from the common replay buffer, where we use $U$ to compute their uncertainties. For computational efficiency, we do not recompute the uncertainty for any states after they have been added to the buffer. This means that for older states, their associated uncertainty values will be stale; however this is not that detrimental since the consequence will be that there may be a slight delay to wait for older, more uncertain states to be pushed out of the buffer before newer, less uncertain states are picked. The staleness can be controlled by changing the maximum capacity of the buffer, $N$, though it is also important to keep the buffer large enough so that the uncertain states get many opportunities to be picked as goals. Due to function approximation, we may need to visit the same goal states over and over again to accumulate enough data to train, which is actually in line with keeping stale uncertainty values.

When picking goals from the goal buffer, we will sample uniformly at random from the top $K$ most uncertain goal states. For simple domains where the uncertainty is accurate and not noisy, it is usually sufficient to use $K=1$ or something very small, and thus always try to reach the most uncertain goal state. However for more complicated domains or much more noisy and inaccurate uncertainty estimates, setting $K$ larger such as $K=100$ results in much more robust behavior. Sampling uniformly from the top $K$ rather than with probability proportional to their uncertainties allows the algorithm to explore a wider variety of goal states in case some of the uncertainty estimates are very inaccurate.

\subsubsection{Computing Uncertainty}

The uncertainty measure $U$ (grey node in Figure \ref{fig:dediagram}) can be something simple, such as count-based bonuses in the tabular setting, or something complex like density models or bayesian models \citep{houthooft2016vime,ostrovski2017count}. In our experiments, we use a simple, learned, uncertainty measure, as we intend to show how directed exploration can take advantage and be more robust to uncertainty and reward bonus-based approaches. We train a network to predict the next state from the current state and action, and then use the prediction loss as the uncertainty. This is also trained from the same minibatches from the common replay buffer.

\subsubsection{On-Policy Mixing}

For each episode, we flip a coin to decide whether we will do directed exploration or just execute the current greedy policy from $A$. The reason we mix in on-policy execution in addition to the off-policy directed exploration samples is because function approximators work much better with on-policy data. Mixing in on-policy data allows for the function approximators to update values for states along the current greedy path and move it towards the optimal policy. The ratio of on-policy vs. directed exploration is a hyperparameter; however we found that $0.5$ is a reasonable value to use in general and did not attempt further optimization.

\section{Robustness Example}

Here, we describe a simple example in which directed exploration is much more robust to the uncertainty estimate than reward bonus approaches. Suppose there are two states $s_1, s_2$ whose uncertainty estimates are close, but $s_1$ is always more uncertain than $s_2$. Then, a reward bonus approach would converge to an exploration policy that always explores $s_1$ and ignores $s_2$. However, with directed exploration, since both $s_1$ and $s_2$ have high uncertainty, they would both be sampled as goal states. Thus directed exploration would end up visiting both of these states.

If the uncertainty estimate is accurate, then it would not be a problem for reward bonus to focus solely on $s_1$, because eventually after visiting $s_1$, its uncertainty estimate would go down and drop below the uncertainty of $s_2$. Then reward bonus would try to visit $s_2$. However if the uncertainty estimate is not accurate, then it is possible that reward bonus would completely miss out on exploring $s_2$, whereas directed exploration is robust to these small relative differences and will end up exploring both.

\section{Tabular Experiment}

\begin{figure}[tb]
    \centering
    \includegraphics[width=0.4\textwidth]{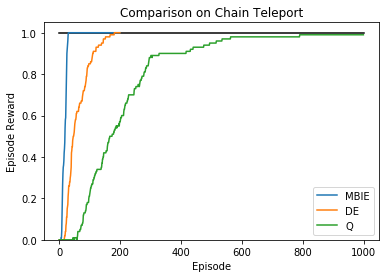}
    \caption{Comparison on gridworld with teleportation. Q-Learning with e-greedy and reward bonus (Q) takes much longer to converge than directed exploration (DE) or MBIE. These results are averaged over 100 runs, and show the evaluation performance when we run the greedy policy.}
    \label{fig:teleport20comparison}
\end{figure}

We first examine the tabular setting where we can compute everything exactly, so we can implement Algorithm \ref{alg:genericdirected} directly. All algorithms are implemented exactly using a table of values and we run value iteration till convergence at every step and so there is no approximation error. 

Here we use a small toy gridworld to illustrate the benefit of committing to reaching states instead of relying on e-greedy. This can come up in settings where a small mistake can be very costly, such as a teleport back to the beginning. The gridworld is a chain of 40 states, where the goal state is at the left end and the initial state is in the middle. Taking actions up or down always teleports you back to the middle, thus taking a random action in the process of reaching the goal can immediately send you back and lose all progress.

We compare 3 algorithms, MBIE \citep{strehlMBIE2008}, Q-learning+bonus (Q), and directed exploration (DE) (Figure \ref{fig:teleport20comparison}). MBIE is a near-optimal model-based algorithm in the tabular setting, and is our best-case baseline after tuning confidence intervals. We use exact count-based reward bonuses as our uncertainty measure. Q-learning+bonus simulates Q-learning with infinite replay and count-based reward bonuses (the same bonuses used for MBIE), where we do not have access to a model and thus cannot learn anything about $(s,a)$ pairs that have never been visited. In such a case, e-greedy exploration is necessary in order to discover new $(s,a)$ pairs not visited yet.

Figure \ref{fig:teleport20comparison} shows, unsurprisingly, that MBIE is the most efficient. Q-learning+bonus (Q) takes a long time to become consistent because there are many runs in which a random exploratory action results in a teleport back to the beginning. Thus trying to reach the goal state at the end of the chain becomes very inefficient. On the other hand, for directed exploration (DE), we only take a random exploratory action once we reach a desired goal state, meaning there is no chance to deviate and fail to reach the goal state due to randomness. This commitment to reaching goal states is another aspect in which directed exploration can be more efficient than reward bonus approaches.

\section{Function Approximation Experiment in a Small Domain}

\begin{figure}[tb]
    \centering
    \includegraphics[width=0.4\textwidth]{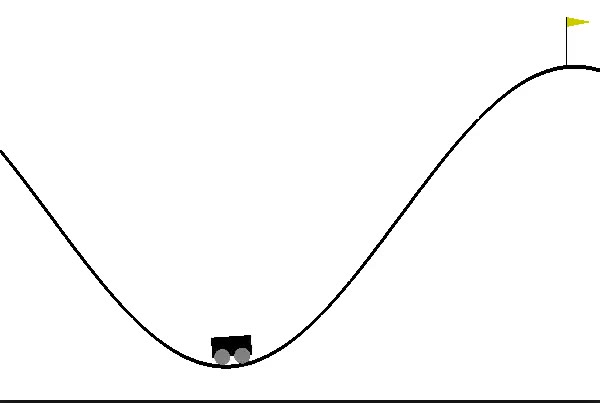}
    \caption{MountainCar Environment \citep{openaigym}}
    \label{fig:mountaincarenv}
\end{figure}

We now move into the function approximation setting by using DQN in the standard Mountain Car domain with discrete actions \citep{openaigym}. We use the double DQN algorithm as the off-policy RL algorithm \citep{van2016deep,baselines} to learn the UVFA with HER, and as the second, regular DQN. To compute the uncertainty of a state, we use the loss from a simple L2 regression task for predicting the next state from the current state and action.

\begin{figure}[tb]
    \centering
    \includegraphics[width=0.4\textwidth]{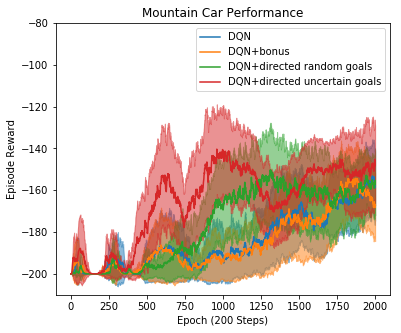}
    \caption{Episode Reward on Mountain Car. Result is over 10 runs, and show the evaluation performance when we run the greedy policy.}
    \label{fig:mountaincarcomparison}
\end{figure}

\begin{figure}[tb]
    \centering
    \includegraphics[width=0.4\textwidth]{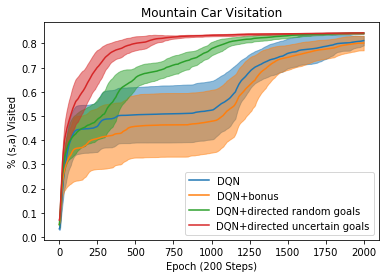}
    \caption{Percent of visited on Mountain Car over a discretized 10x10 grid. Result is over 10 runs.}
    \label{fig:mountaincarvisit}
\end{figure}

We compare 4 algorithms: DQN, DQN+bonus, DQN+directed with sampling uniformly random goals, and DQN+directed picking the most uncertain goals. For DQN and DQN+bonus, as well as the second regular DQNs present in the directed exploration algorithms, we use e-greedy exploration with a fixed $\epsilon = 0.1$. This is because even though DQN+bonus uses a reward bonus, it cannot assign reward bonuses to states that it has never seen before (bonuses are augmented to minibatches sampled from the replay buffer), and thus must still rely on some form of randomness to discover new states. Directed exploration does not need randomness when trying to reach goal states; instead it takes a step of random action after reaching a goal or reaching the timelimit $D$.

From Figure \ref{fig:mountaincarcomparison}, we see that directed exploration using uncertainty (red) is the fastest at learning. Directed exploration with random goals (green) comes in second. Regular DQN (blue) and DQN with reward bonus (orange) seem to tie and perform the worst. These results indicate that directed exploration by itself, even just using random goals and not using uncertainty, is already better at finding the sparse reward. But the best results come from using the uncertainty to drive directed exploration towards the most uncertain states. This shows that the uncertainty, based on prediction loss, is in fact informative and useful for exploration. Reward bonus does not seem to be able to take advantage of this uncertainty at all.

To gather more insight, Figures \ref{fig:mountaincarvisit} shows how much of the state space (after discretization) has been visited at least once. We see a much clearer difference in terms of how the algorithms are exploring. Both versions of directed exploration are much more efficient at covering the state space, whereas reward bonus seems to not be able to explore more than regular DQN. But the uncertainty is clearly important, as directed exploration using uncertainty to pick goals is significantly faster at exploration than when picking goals randomly. These results directly correlate with the actual performance of these algorithms in Figure \ref{fig:mountaincarcomparison}.

The uncertainty based on prediction loss that we use is a very simple and natural approach that has been explored in prior works \citep{burda2018large}. It is not a sophisticated, state-of-the-art measure of uncertainty, and thus is more noisy and less accurate. The reward bonus approach struggles and is not able to explore using this uncertainty. However, directed exploration is much more robust and is still able to take advantage of such a measure of uncertainty to explore extremely efficiently.

\subsection{Experimental Details}

Hyperparameters used for DQN were two hidden layers of size 64 with RELU activations. We used the Adam optimizer with a learning rate of 0.0001, target network is updated every 1000 steps. Prioritized replay is used with $\alpha = 0.4$, and $\beta = 1.0$. We used double DQN with a huber loss function. For the UVFA, we had a separate DQN with the same architecture, except we used a learning rate of 0.001 and the target is updated every 30 steps, and we used uniform replay. We found that the UVFA network could sustain a high learning rate and therefore converge much faster, unlike the regular DQN network which sometimes diverged with higher learning rates. For e-greedy, we used a fixed $\epsilon = 0.1$.

For reward bonus, we found that normalizing the uncertainty using running exponential averages of minimum and maximums, and then scaling by the reciprocal of the square root of the number of timesteps helps stablize performance.

For directed exploration, we set $K=1$, meaning we always pick the most uncertain state from the goal buffer. We set the maxmimum length of a directed exploration attempt, $D=50$, which is a quarter of the maximum episode length. The goal buffer has a capacity of $10000$.

\section{Function Approximation Experiment in a Large Domain}

\begin{figure}[tb]
    \centering
    \includegraphics[width=0.4\textwidth]{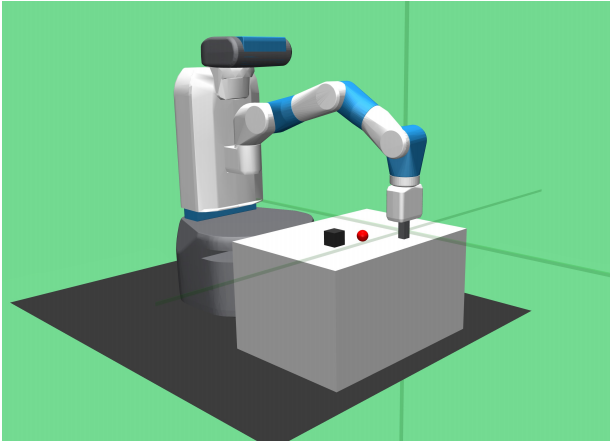}
    \caption{FetchPush Domain, image from \citep{plappert2018multi}.}
    \label{fig:fetchpushenv}
\end{figure}

We implemented directed exploration in the FetchPush domain \citep{andrychowicz2017hindsight, plappert2018multi} (Figure \ref{fig:fetchpushenv}). This domain consists of a robotic arm in the middle of a table. The objective is to push the box to a desired goal position (3 dimensional) on the table. The initial position of the box is random. In the original formulation, the goal position is known and also randomized every episode, which is done in order to perform multi-goal learning. We have modified the domain to a fixed but unknown goal position to turn it into an ordinary MDP environment. For directed exploration, we also use object positions as goal states to be consistent with the domain. We rely on DDPG as the off-policy RL algorithm since actions are continuous \cite{plappert2018multi}. The uncertainty we use is the mean squared loss from predicting the next object position from the current state and action.

\begin{figure}[tb]
    \centering
    \includegraphics[width=0.4\textwidth]{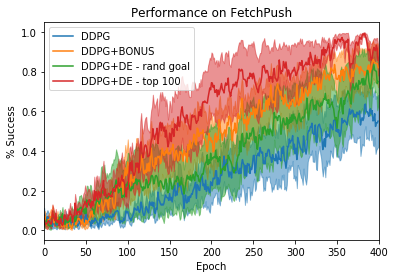}
    \caption{Percentage of success of reaching the true goal on FetchPush when comparing regular DDPG, DDPG and reward bonus, and DDPG with directed exploration both with picking random goals and top 100 most uncertain goals. Results are over 5 runs.}
    \label{fig:fetchpushcomparison}
\end{figure}

\begin{figure}[tb]
    \centering
    \includegraphics[width=0.4\textwidth]{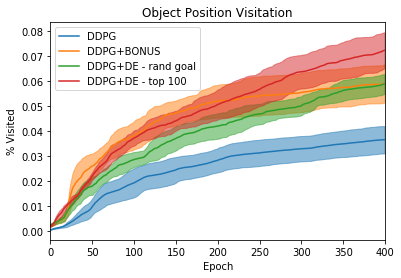}
    \caption{Percentage of visitation of possible discretized goal states on FetchPush when comparing regular DDPG, DDPG and reward bonus, and DDPG with directed exploration both with picking random goals and top 100 most uncertain goals. Results are over 5 runs.}
    \label{fig:fetchpushvisit}
\end{figure}

Figure \ref{fig:fetchpushcomparison} compares the performance of regular DDPG (blue), DDPG with reward bonus (orange), DDPG with directed exploration but samples uniformly random goals from the goal buffer (green), and DDPG with directed exploration but samples the top 100 most uncertain goals from the goal buffer (red). The performance is measured by evaluating the learned policy on 10 new episodes and calculating the percentage that successfully reach the true fixed goal \citep{andrychowicz2017hindsight}.

We see from Figure \ref{fig:fetchpushcomparison} that directed exploration with sampling from the top 100 most uncertain goals (red) is able to learn the fastest compared to the others. Notably, it is more efficient than directed exploration with sampling random goals from the goal buffer (green). This indicates that the uncertainty we are using is actually meaningful, and driving exploration towards uncertain goals is more efficient than random goals. The reward bonus approach using the same uncertainty (orange) is still better than regular DDPG (blue), further confirming that the uncertainty is informative for exploration, but is not as efficient as directed exploration. Furthermore, for the reward bonus approach, we had to normalize the bonus by keeping a running minimum and maximum, and rescale it to decay as the reciprical of the number of timesteps in order to obtain the current performance; using the raw mean squared prediction loss simply results in identical performance to regular DDPG. On the other hand, for directed exploration, we used the raw losses and stored them in the goal buffer to use for sampling. These results show that directed exploration is able to much better take advantage and extract out meaningful information from the uncertainty estimate, making it much more robust than the reward bonus approach.

To develop more insight, we have also tracked how many different object positions have been explored in Figure \ref{fig:fetchpushvisit}. We discretized the possible object positions and tracked what percentage of total positions have been visited at least once. The results show that directed exploration using the uncertainty (red) is able to explore more than directed exploration with random goals (green), which confirms the effectiveness of using unertainty to guide exploration. The reward bonus approach (orange) initially falls short of directed exploration with uncertainty, before slightly overtaking it in the middle, but eventually still falls short. This is because once the reward bonus approach has found the sparse reward, then it will gradually stop exploration and start exploitation. It seems that even though in the middle, reward bonus is able to match and slightly exceed the amount of exploration that directed exploration is able to achieve, it is slower to start, and thus slower at first encountering the sparse reward. It would seem that the non-stationary reward bonuses slow down convergence to the optimal policy, whereas directed exploration avoids non-stationarity and is able to converge faster.

\subsection{Experimental Details}

We use the DDPG and HER implementations from \cite{plappert2018multi}, along with the same hyperparameters. Due to technical limitations, we did not run 19 parallel threads, but only 1. For directed exploration, we use two DDPG agents; one is goal-conditioned and is trained using HER; the second is regular unconditioned DDPG. The performance is evaluated solely using the second regular DDPG.

For predicting the next object position from the current state and action, we use a neural network that takes as input the current state and action, passes it through 2 hidden layers of with 256 units each and RELU activations, and a final linear output layer that predicts the next object position (3 dimensional). This is trained using Adam with a learning rate of $0.001$ and a batch size of 256, in tandem with DDPG.

For reward bonus, we keep a running exponential average of the minimum and maximum uncertainty from this mean squared loss, and rescale the raw loss to be between 0 and 1, before dividing it by the square root of the number of minibatch updates so far to better emulate a count-based bonus. This turns out to be much more effective than using the raw mean squared loss directly.

For directed exploration, the goal buffer has capacity 50000. We set $K=100$, meaning we sample uniformly from the top 100 most uncertain states in the goal buffer. We set the maxmimum length of a directed exploration attempt, $D$, to be $25$, half of the episode length. When we use the goal-conditioned DDPG, we use the greedy policy and disable any action noise.

\section{Discussion of Results}

The tabular setting showed how committing to reaching goal states with high uncertainty was much more efficient than the combination of e-greedy and reward bonus. However, if e-greedy could be avoided, like using a model-based method such as MBIE, then it is possible to achieve more efficient exploration. But in the function approximation case, learning a good dynamics model is a big challenge, and the current most successful methods actually combine model-based and model-free, resulting in still using some form of randomization in the actions \citep{nagabandi2018neural}. Thus, commitment in directed exploration may still result in increased efficiency for exploration. However, commitment only has the largest impact if small mistakes are extremely costly, and this is not always the case for many domains. If small mistakes are not that bad, then commitment wil not result in a drastic increase in efficiency for exploration.

The function approximation experiments showed how directed exploration is more efficient at exploration, but more importantly, can take advantage of a simple uncertainty measure to explore even better. Before we discuss the uncertainty, we will first discuss directed exploration itself. The results on Mountain Car when sampling goal states at random do show that directed exploration itself is already faster at exploration. However the results on FetchPush show that sampling random goal states is not as good at exploration as reward bonus. This difference highlights one of the biggest challenges for directed exploration: learning the goal-conditioned policy $\pi(s,g)$. In Mountain Car, due to the simplicity of the domain, we are able to learn $\pi(s,g)$ very efficiently, and thus directed exploration is able to start exploring efficiently very early on. On FetchPush, it takes much longer (Figure \ref{fig:fetchpushvisit}), and so directed exploration with random goals is not able to explore a lot until later on in training. We hypothesize that if we are able to train a $\pi(s,g)$ faster, whether using a different architecture or improving the training process, then directed exploration may become much more efficient at exploration.

We will now discuss directed exploration with uncertainty. For both Mountain Car and FetchPush, using the uncertainty to sample goals greatly increased the efficiency of exploration over just using random goals. This indicates that the uncertainty we used is in fact informative about what parts of the state space we need to explore, even though our uncertainty is simply the prediction loss from predicting the dynamics. However when we compare performance to reward bonus, we see that in Mountain Car, reward bonus is not able to take advantage of the uncertainty at all. In FetchPush, the reward bonus is able to take some advantage of the uncertainty and improve performance, but not as much as directed exploration. Even though in FetchPush, the goal-conditioned policy $\pi(s,g)$ is being learned quite slowly, it seems that trying to reach those goal states with high uncertainty makes a significant difference to exploration efficiency. These results show that directed exploration is robust to the uncertainty measure and can better take advantage to explore more than reward bonus. Reward bonus seems to be much slower to update and propagate the uncertainty.

There is an additional computation cost for directed exploration since we need to maintain two off-policy RL algorithms. But there could be ways to reduce the cost, such as potentially sharing parts of the network between the two algorithms. Alternatively, the two algorithms can easily be updated in parallel, taking advantage of multithreading to speed up computation.

\section{Related Work}

Recently, Uber introduced the Go-Explore algorithm \footnote{\url{https://eng.uber.com/go-explore/}}, which has achieved incredible results on two hard exploration Atari games: Montezuma's Revenge and Pitfall. One of the key components of their algorithm is to use directed exploration to revisit states that looked promising for further exploration. However their approach differs from ours as they are not learning a general goal-conditioned policy to figure out how to reach states, but they are replaying past trajectories due to being in a deterministic environment. Our approach is general and fully support stochastic environments. Furthermore their algorithm also has many other components put together and it is not clear the effect of directed exploration itself.

Directed exploration can be considered a particular instantiation of goal generation from higher level policies in hierarchical reinforcement learning. Prior work has focused on uniformly random or expert guided goal generation for learning sub task and task structure, but we look into uncertainty-based goal generation explicitly for better exploration \citep{Held2018AutomaticGG,NIPS2018_7591}. Another difference is that our directed exploration component is completely separated from the existing off-policy algorithm, and it only contributes samples to the common replay buffer. This allows our component to be more modular, and more easily added to existing off-policy algorithms.

\section{Acknowledgements}

We would like to thank Microsoft and Google for helping make this research possible.

\section{Conclusion}

We presented a general directed exploration algorithm for RL with function approximation, and showed how it can be more efficient at exploration using the same measure of uncertainty as reward bonus approaches. This is due to directed exploration committing to reach states, and being more robust to noisy or inaccurate uncertainty estimates.

For future work, there is still a lot more potential to try directed exploration on even larger and more complex domains, where learning goal-conditioned policies becomes much more challenging. It would be interesting to see whether simple uncertainty estimates that have not worked with reward bonus approaches would be sufficient for directed exploration to explore efficiently.

\bibliographystyle{plainnat}
%\bibliography{main}

\begin{thebibliography}{20}
\providecommand{\natexlab}[1]{#1}
\providecommand{\url}[1]{\texttt{#1}}
\expandafter\ifx\csname urlstyle\endcsname\relax
  \providecommand{\doi}[1]{doi: #1}\else
  \providecommand{\doi}{doi: \begingroup \urlstyle{rm}\Url}\fi

\bibitem[Andrychowicz et~al.(2017)Andrychowicz, Wolski, Ray, Schneider, Fong,
  Welinder, McGrew, Tobin, Abbeel, and Zaremba]{andrychowicz2017hindsight}
Marcin Andrychowicz, Filip Wolski, Alex Ray, Jonas Schneider, Rachel Fong,
  Peter Welinder, Bob McGrew, Josh Tobin, OpenAI~Pieter Abbeel, and Wojciech
  Zaremba.
\newblock Hindsight experience replay.
\newblock In \emph{Advances in Neural Information Processing Systems}, pages
  5048--5058, 2017.

\bibitem[Brockman et~al.(2016)Brockman, Cheung, Pettersson, Schneider,
  Schulman, Tang, and Zaremba]{openaigym}
Greg Brockman, Vicki Cheung, Ludwig Pettersson, Jonas Schneider, John Schulman,
  Jie Tang, and Wojciech Zaremba.
\newblock Openai gym, 2016.

\bibitem[Burda et~al.(2018)Burda, Edwards, Pathak, Storkey, Darrell, and
  Efros]{burda2018large}
Yuri Burda, Harri Edwards, Deepak Pathak, Amos Storkey, Trevor Darrell, and
  Alexei~A Efros.
\newblock Large-scale study of curiosity-driven learning.
\newblock \emph{arXiv preprint arXiv:1808.04355}, 2018.

\bibitem[Dhariwal et~al.(2017)Dhariwal, Hesse, Klimov, Nichol, Plappert,
  Radford, Schulman, Sidor, Wu, and Zhokhov]{baselines}
Prafulla Dhariwal, Christopher Hesse, Oleg Klimov, Alex Nichol, Matthias
  Plappert, Alec Radford, John Schulman, Szymon Sidor, Yuhuai Wu, and Peter
  Zhokhov.
\newblock Openai baselines.
\newblock \url{https://github.com/openai/baselines}, 2017.

\bibitem[Fortunato et~al.(2017)Fortunato, Azar, Piot, Menick, Osband, Graves,
  Mnih, Munos, Hassabis, Pietquin, et~al.]{fortunato2017noisy}
Meire Fortunato, Mohammad~Gheshlaghi Azar, Bilal Piot, Jacob Menick, Ian
  Osband, Alex Graves, Vlad Mnih, Remi Munos, Demis Hassabis, Olivier Pietquin,
  et~al.
\newblock Noisy networks for exploration.
\newblock \emph{arXiv preprint arXiv:1706.10295}, 2017.

\bibitem[Held et~al.(2018)Held, Geng, Florensa, and
  Abbeel]{Held2018AutomaticGG}
David Held, Xinyang Geng, Carlos Florensa, and Pieter Abbeel.
\newblock Automatic goal generation for reinforcement learning agents.
\newblock In \emph{ICML}, 2018.

\bibitem[Houthooft et~al.(2016)Houthooft, Chen, Duan, Schulman, De~Turck, and
  Abbeel]{houthooft2016vime}
Rein Houthooft, Xi~Chen, Yan Duan, John Schulman, Filip De~Turck, and Pieter
  Abbeel.
\newblock Vime: Variational information maximizing exploration.
\newblock In \emph{Advances in Neural Information Processing Systems}, pages
  1109--1117, 2016.

\bibitem[Lillicrap et~al.(2015)Lillicrap, Hunt, Pritzel, Heess, Erez, Tassa,
  Silver, and Wierstra]{lillicrap2015continuous}
Timothy~P Lillicrap, Jonathan~J Hunt, Alexander Pritzel, Nicolas Heess, Tom
  Erez, Yuval Tassa, David Silver, and Daan Wierstra.
\newblock Continuous control with deep reinforcement learning.
\newblock \emph{arXiv preprint arXiv:1509.02971}, 2015.

\bibitem[Mnih et~al.(2013)Mnih, Kavukcuoglu, Silver, Graves, Antonoglou,
  Wierstra, and Riedmiller]{mnih2013playing}
Volodymyr Mnih, Koray Kavukcuoglu, David Silver, Alex Graves, Ioannis
  Antonoglou, Daan Wierstra, and Martin Riedmiller.
\newblock Playing atari with deep reinforcement learning.
\newblock In \emph{NIPS Deep Learning Workshop}. 2013.

\bibitem[Mnih et~al.(2016)Mnih, Badia, Mirza, Graves, Lillicrap, Harley,
  Silver, and Kavukcuoglu]{mnih2016asynchronous}
Volodymyr Mnih, Adria~Puigdomenech Badia, Mehdi Mirza, Alex Graves, Timothy
  Lillicrap, Tim Harley, David Silver, and Koray Kavukcuoglu.
\newblock Asynchronous methods for deep reinforcement learning.
\newblock In \emph{International conference on machine learning}, pages
  1928--1937, 2016.

\bibitem[Nachum et~al.(2018)Nachum, Gu, Lee, and Levine]{NIPS2018_7591}
Ofir Nachum, Shixiang~(Shane) Gu, Honglak Lee, and Sergey Levine.
\newblock Data-efficient hierarchical reinforcement learning.
\newblock In S.~Bengio, H.~Wallach, H.~Larochelle, K.~Grauman, N.~Cesa-Bianchi,
  and R.~Garnett, editors, \emph{Advances in Neural Information Processing
  Systems 31}, pages 3307--3317. Curran Associates, Inc., 2018.

\bibitem[Nagabandi et~al.(2018)Nagabandi, Kahn, Fearing, and
  Levine]{nagabandi2018neural}
Anusha Nagabandi, Gregory Kahn, Ronald~S Fearing, and Sergey Levine.
\newblock Neural network dynamics for model-based deep reinforcement learning
  with model-free fine-tuning.
\newblock In \emph{2018 IEEE International Conference on Robotics and
  Automation (ICRA)}, pages 7559--7566. IEEE, 2018.

\bibitem[Osband et~al.(2016)Osband, Blundell, Pritzel, and
  Van~Roy]{osband2016deep}
Ian Osband, Charles Blundell, Alexander Pritzel, and Benjamin Van~Roy.
\newblock Deep exploration via bootstrapped dqn.
\newblock In \emph{Advances in neural information processing systems}, pages
  4026--4034, 2016.

\bibitem[Ostrovski et~al.(2017)Ostrovski, Bellemare, Oord, and
  Munos]{ostrovski2017count}
Georg Ostrovski, Marc~G Bellemare, Aaron van~den Oord, and R{\'e}mi Munos.
\newblock Count-based exploration with neural density models.
\newblock \emph{arXiv preprint arXiv:1703.01310}, 2017.

\bibitem[Plappert et~al.(2018)Plappert, Andrychowicz, Ray, McGrew, Baker,
  Powell, Schneider, Tobin, Chociej, Welinder, et~al.]{plappert2018multi}
Matthias Plappert, Marcin Andrychowicz, Alex Ray, Bob McGrew, Bowen Baker,
  Glenn Powell, Jonas Schneider, Josh Tobin, Maciek Chociej, Peter Welinder,
  et~al.
\newblock Multi-goal reinforcement learning: Challenging robotics environments
  and request for research.
\newblock \emph{arXiv preprint arXiv:1802.09464}, 2018.

\bibitem[Schaul et~al.(2015)Schaul, Horgan, Gregor, and
  Silver]{schaul2015universal}
Tom Schaul, Daniel Horgan, Karol Gregor, and David Silver.
\newblock Universal value function approximators.
\newblock In \emph{International Conference on Machine Learning}, pages
  1312--1320, 2015.

\bibitem[Schulman et~al.(2017)Schulman, Wolski, Dhariwal, Radford, and
  Klimov]{schulman2017proximal}
John Schulman, Filip Wolski, Prafulla Dhariwal, Alec Radford, and Oleg Klimov.
\newblock Proximal policy optimization algorithms.
\newblock \emph{arXiv preprint arXiv:1707.06347}, 2017.

\bibitem[Strehl and Littman(2008)]{strehlMBIE2008}
Alexander~L. Strehl and Michael~L. Littman.
\newblock {An analysis of model-based Interval Estimation for Markov Decision
  Processes}.
\newblock \emph{Journal of Computer and System Sciences}, 74\penalty0
  (8):\penalty0 1309--1331, 2008.

\bibitem[Tang et~al.(2017)Tang, Houthooft, Foote, Stooke, Chen, Duan, Schulman,
  DeTurck, and Abbeel]{tang2017exploration}
Haoran Tang, Rein Houthooft, Davis Foote, Adam Stooke, OpenAI~Xi Chen, Yan
  Duan, John Schulman, Filip DeTurck, and Pieter Abbeel.
\newblock \# exploration: A study of count-based exploration for deep
  reinforcement learning.
\newblock In \emph{Advances in Neural Information Processing Systems}, pages
  2753--2762, 2017.

\bibitem[Van~Hasselt et~al.(2016)Van~Hasselt, Guez, and Silver]{van2016deep}
Hado Van~Hasselt, Arthur Guez, and David Silver.
\newblock Deep reinforcement learning with double q-learning.
\newblock In \emph{AAAI}, volume~2, page~5. Phoenix, AZ, 2016.

\end{thebibliography}

\end{document}